\documentclass[letterpaper, 10 pt, conference, dvipsnames]{ieeeconf}

\usepackage{multirow}
\usepackage{makecell}
\usepackage{bm}
\usepackage{amsmath}
\usepackage{amssymb}
\usepackage{booktabs}
\usepackage{graphicx}
\usepackage{tabularx}
\usepackage{ragged2e}
\usepackage{colortbl}
\usepackage{algorithm}
\usepackage{algorithmicx}
\usepackage{algpseudocode}
\usepackage{caption}
\usepackage{subcaption}
\usepackage[table]{xcolor}
\PassOptionsToPackage{dvipsnames}{xcolor}
\usepackage{pifont}
\let\labelindent\relax
\usepackage{enumitem}
\usepackage{tikz}
\usepackage{cite}
\usetikzlibrary{positioning, shapes, arrows.meta, calc}
\usepackage{csquotes}
\usepackage{hyperref}

\definecolor{DarkBlue}{RGB}{31,78,121}
\definecolor{DarkGreen}{RGB}{46,125,50}

\IEEEoverridecommandlockouts
\title{\LARGE \bf
BEACON: Language-Conditioned Navigation Affordance Prediction\\under Occlusion
}

\author{Xinyu Gao, Gang Chen$^{\dag}$, Javier Alonso-Mora
\thanks{$\dag$ Corresponding author}
\thanks{The authors are with the Department of Cognitive Robotics (CoR), Delft
University of Technology
{\tt\small x.gao-14@student.tudelft.nl; \{g.chen-5; j.alonsomora\}@tudelft.nl} %
}
}

\makeatletter
\let\@oldmaketitle\@maketitle
\renewcommand{\@maketitle}{%
    \@oldmaketitle%
    \vspace*{-1mm}
    \centering
    \includegraphics[width=\textwidth]{new_imgs/teeaser_v7.pdf}
    \vspace*{-5mm}
    \captionof{figure}{BEACON predicts an ego-centric Bird's-Eye View (BEV) affordance heatmap for language-conditioned local navigation, which is better suited to occluded targets than the state-of-the-art image-space grounding method.}
    \label{fig:intro_teaser}
    \vspace*{-2mm}
}
\makeatother

\begin{document}

\maketitle
\setcounter{figure}{1} %
\thispagestyle{empty}
\pagestyle{empty}

\begin{abstract}
Language-conditioned local navigation requires a robot to infer a nearby traversable target location from its current observation and an open-vocabulary, relational instruction. %
Existing vision-language spatial grounding methods usually %
rely on vision–language models (VLMs) to reason in image space, producing 2D predictions tied to visible pixels.
As a result, they struggle to infer target locations in occluded regions, typically caused by furniture or moving humans. 
To address this issue, %
we propose BEACON, which predicts an ego-centric Bird’s-Eye View (BEV) affordance heatmap over a bounded local region including occluded areas.
Given an instruction and surround-view RGB-D observations from four directions around the robot, BEACON predicts the BEV heatmap by injecting spatial cues into a VLM and fusing the VLM’s output with depth-derived BEV features.
Using an occlusion-aware dataset built in the Habitat simulator,
we conduct detailed experimental analysis
  to validate both our BEV space formulation and the design choices of each module. 
Our method improves the accuracy averaged across geodesic thresholds by 22.74 percentage points over the state-of-the-art image-space baseline on the validation subset with occluded target locations. Our project page is: \href{https://xin-yu-gao.github.io/beacon}{https://xin-yu-gao.github.io/beacon}.

\end{abstract}

\section{Introduction}
\label{sec:introduction}

Language-conditioned local navigation requires a robot to decide where to go from natural instructions that describe a nearby traversable target location through ego-centric directions, landmarks, or scene layout (e.g., “go behind the table,” “turn left and move forward,” or “go down the hallway”). Unlike tasks that can be addressed by detecting a single object instance, it requires spatial understanding and grounding to a precise traversable target location. In cluttered indoor environments, the target location can be hard to ground from the current observations due to occlusions caused by furniture or people, yet in many cases the robot still needs to choose a feasible local target. %
This setting requires the robot to infer, from language and current observations, a local target location in its ego-centric frame that is traversable, even when the target is occluded.

Recent vision-language spatial grounding methods~\cite{yuan2025robopoint, zhouroborefer, liu2025spatialcot} use vision-language models (VLMs) to map observations and instructions to spatial targets and represent the closest existing setup to our problem. These models typically produce image-space point predictions and demonstrate strong open-vocabulary spatial understanding across diverse scenes. However, because image-space outputs are tied to what is directly visible in a particular view, these models struggle to predict target locations under occlusion in the current observations. 
Occlusion-aware spatial perception for robots has been studied~\cite{song2017semantic, reed2024scenesense}, but generally not as a language-conditioned local target prediction problem.
Meanwhile, ego-centric Bird’s-Eye-View (BEV) representations have proven effective for producing ground-plane outputs under occlusion~\cite{zhang2026vision}, and recent works show that injecting BEV feature or 3D cues to VLM can improve its performance on ego-centric tasks~\cite{zhu2025llava, qu2025spatialvla, shao2024lmdrive}. Together, these developments motivate combining VLM-based spatial grounding with ego-centric 3D cues and a robot-centric BEV output for local navigation target prediction under occlusion.

In this work, we propose \textbf{BEACON}, a BEV-enhanced affordance prediction model for language-conditioned local navigation under occlusion, shown in Figure~\ref{fig:intro_teaser}. Given single-timestep surround-view RGB-D observations and a natural language instruction, BEACON predicts an ego-centric BEV affordance heatmap over nearby ground locations. 
Here, \emph{affordance} denotes the score indicating how suitable each location is as a local navigation target.
BEACON combines an Ego-Aligned Vision-Language Model for instruction-conditioned ego-centric scene understanding with a Geometry-Aware Bird's-Eye View Encoder that provides metric spatial structure from RGB-D observations, allowing the model to infer traversable local targets even when they are occluded in the current views. %

In detail, our main contributions are as follows:
We propose a single-timestep ego-centric BEV navigation affordance prediction method that grounds open-vocabulary instructions into a local BEV affordance heatmap, making it better suited to occluded targets than image-space spatial grounding.
We propose an Ego-Aligned VLM that incorporates 3D positional cues to improve language-conditioned target prediction, together with a BEV-space affordance formulation trained with explicit negatives over non-traversable regions to encourage structural validity.
Systematic experiments on an occlusion-aware dataset in the Habitat~\cite{puig2024habitat3} simulator show consistent gains over zero-shot image-space baselines and trained architecture variants under occlusion, validating the role of each design component.

\section{Related Work}
\subsection{Vision-Language Spatial Grounding in Robotics}
The most relevant line of work to our problem is vision-language spatial grounding in robotics, where models map observations and instructions to spatial intermediate outputs that can be consumed by downstream planners. A non-VLM method
~\cite{kim2024lingo} struggles with non-object descriptions due to its object-centric design.
VLM-based methods typically predict one or a few 2D coordinates as target points, often projecting them to 3D using depth for execution. RoboPoint~\cite{yuan2025robopoint} shows that instruction-tuning with synthetic object-reference and free-space reference data enables a general VLM to output image-space points satisfying spatial relations, and demonstrates downstream use in navigation and manipulation. Follow-up directions improve spatial capability via richer supervision~\cite{shao2025more}, explicit reasoning~\cite{zhouroborefer, liu2025spatialcot}, or additional geometric cues~\cite{zhouroborefer}, often explicitly considering navigation as a downstream use~\cite{yuan2025robopoint, song2025robospatial, tang2025roboafford, hao2025roboafford++, zhouroborefer, liu2025spatialcot, shao2025more}.

These approaches are effective as general-purpose spatial interfaces because they leverage web-scale pretrained visual semantics and language reasoning. However, robot navigation in cluttered indoor scenes often involves occlusions and indirect cues, where the instruction may imply a landmark or target location behind people or structures. Many existing formulations express outputs in image coordinates or prioritize directly observable evidence during inference and evaluation, and they typically do not explicitly target robot-centric local goal inference under occlusions or enforce structural feasibility (e.g., avoiding walls) for local navigation targets. 
Our work focuses on this navigation-centric regime and leverages 3D cues with a robot-centric spatial representation to predict targets under occlusion while promoting traversability.

\subsection{VLMs with Local 3D or Ego-Centric Multi-View Inputs}
Robots often operate with richer geometric observations than a single RGB image, such as depth or multiple ego-centric views, motivating efforts to extend vision-language models with local 3D or ego-centric multi-view inputs for improved spatial understanding.
Some approaches inject 3D cues directly into the 2D vision tokens~\cite{zhu2025llava, qu2025spatialvla, cheng20253d}, whereas others introduce a separate depth or 3D branch~\cite{hong20233d, huang2024embodied, huang2025leo, zhouroborefer, cai2025spatialbot, cheng2024spatialrgpt}.
While they demonstrate strong 3D understanding across captioning, question answering, and grounding tasks, these models are not typically designed to output robot-centric local navigation targets, nor are they commonly evaluated in tasks where the referred target is occluded in the current view.
Recent work also explores ego-centric multi-view spatial reasoning in vision-language models, but remains focused on reasoning-oriented tasks such as question answering rather than local navigation target prediction~\cite{gholami2025spatial}.

\subsection{BEV Representations and VLM Alignment}
BEV representations provide a geometry-centric interface that inherently preserves metric spatial structure and are widely used in occlusion-heavy perception settings~\cite{philion2020lift, li2024bevformer, li2023fb, liu2023bevfusion}, mainly in the self-driving domain. BEV-based methods usually convert visual features to the ground plane using depth or point clouds to produce dense top-down features for various tasks (e.g., detection and segmentation), providing a natural spatial basis for modeling targets behind occlusions while respecting local traversability constraints. In natural language-conditioned tasks, recent advances in self-driving typically provide BEV features as input to the language model, sometimes compressing BEV information into a small set of tokens through adapter modules like~\cite{li2023blip} and not passing raw images to the downstream language model~\cite{shao2024lmdrive, winter2025bevdriver, liu2025drivepi}. While effective for driving objectives, this design may obscure fine-grained spatial structure that is important for precise local goal selection in cluttered indoor environments and may reduce the benefit of web-scale knowledge priors from pretrained vision–language models.

Motivated by these gaps, BEACON aims to retain raw image inputs for VLM-based scene understanding while using depth-derived robot-centric BEV features to preserve local geometry. It then combines language understanding with dense spatial representation to predict instruction-consistent navigation affordance under occlusion while respecting local traversability constraints.

\begin{figure}[t]
  \centering
  \includegraphics[width=\columnwidth]{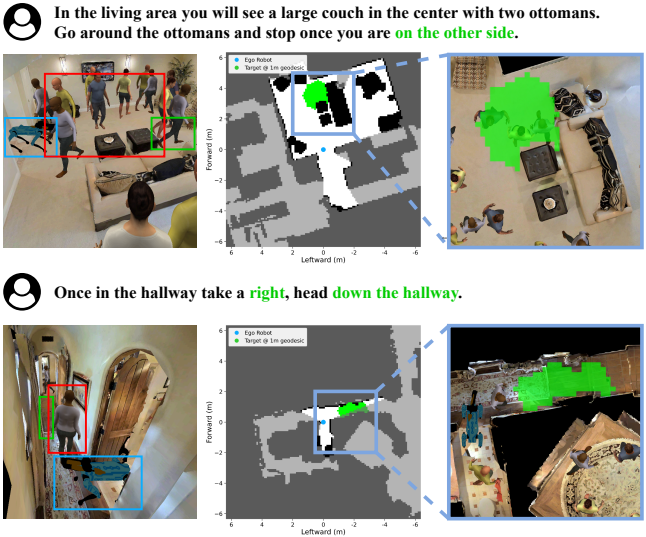}
\caption{Examples of language-conditioned local navigation under occlusion. The \textcolor[HTML]{0AADFF}{blue boxes} mark the robot, the \textcolor{red}{red boxes} highlight humans and objects that cause occlusions, and the \textcolor{green}{green boxes} indicate target regions.}
  \label{fig:task_overview}
  \vspace{-4mm}
\end{figure}

\begin{figure*}[t]
  \centering
  \includegraphics[width=\textwidth]{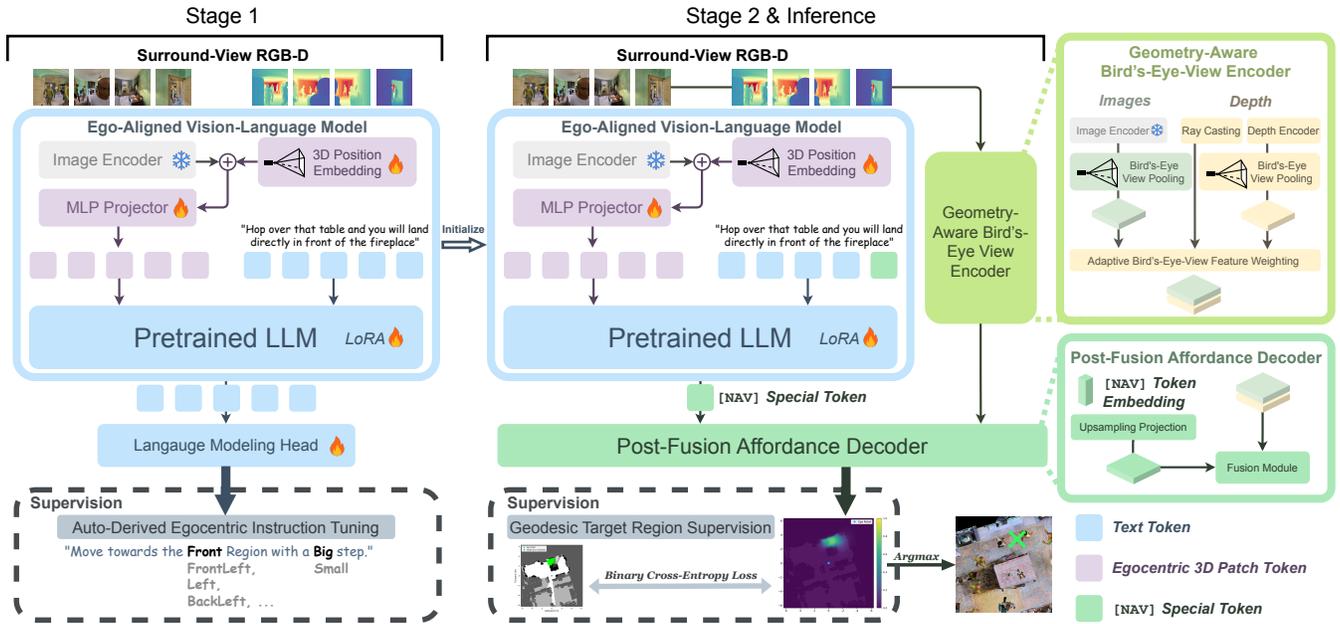}
\caption{BEACON overview. Stage 1 performs auto-derived ego-centric instruction tuning with ego-centric 3D position encoding to train the Ego-Aligned VLM. Stage 2 initializes the Ego-Aligned VLM weights from Stage 1, combines the resulting instruction-conditioned output with Geometry-Aware BEV features, and predicts an ego-centric BEV navigation affordance heatmap via a Post-Fusion Affordance Decoder. The two stages use different supervision signals, and inference selects the navigation target by taking the argmax.}

  \label{fig:pipeline}
   \vspace{-3mm}
\end{figure*}

\section{Problem Formulation}
\label{sec:problem_formulation}

Given four surround RGB-D views $o$ and a human language instruction $x$, we aim to infer an ego-centric local navigation target that matches the instruction and lies in traversable free space. This setting requires grounding open-vocabulary instructions, expressed through landmarks, scene structure, or ego-centric directions, to a local destination rather than detecting a single visible object. The intended target may be partially or fully occluded by static structures and/or transient obstacles while still lying within a bounded local area that does not require exploration. Figure~\ref{fig:task_overview} illustrates this setting: the left panel shows the task setup under occlusion (isometric view), the middle panel shows the ego-centric top-down local grid with a target region visualization, and the right panel overlays the same target region on a simulator top-down view.

We formulate local target inference as dense prediction in the ego-centric BEV space. The model outputs an ego-centric BEV navigation affordance map $\hat{A}$, where higher values indicate more likely instruction-relevant target locations, and a single target point for evaluation is obtained by taking the argmax of $\hat{A}$ and mapping the selected BEV cell to its metric location. Formally, we learn a model $f_{\theta}$ that maps observation and instruction to the affordance map:
\begin{equation}
f_{\theta}(o,x)\rightarrow \hat{A}.
\end{equation}
During training, $\hat{A}$ is supervised with a target region mask around the annotated target point, as defined in Section~\ref{sec:region_supervision}. Dynamic obstacle avoidance and social navigation behaviors (e.g., yielding to pedestrians) are outside the scope.

\section{Methodology}

Figure~\ref{fig:pipeline} summarizes BEACON, which consists of two stages. Stage~1 adapts a pretrained \textbf{Ego-Aligned VLM} (Section~\ref{sec:vlm_3d}) for ego-centric scene understanding from surround-view observations and natural language instructions. To support this adaptation, we incorporate ego-centric 3D position encoding and perform auto-derived ego-centric instruction tuning, so that the model better interprets spatial language in the agent frame under the surround-view setting. Stage~2 then initializes from the Stage~1 Ego-Aligned VLM weights and builds the full navigation affordance predictor by combining the instruction-conditioned VLM output with a \textbf{Geometry-Aware BEV Encoder} (Section~\ref{sec:bev_path}) and a \textbf{Post-Fusion Affordance Decoder} (Section~\ref{sec:post_fusion_decoder}). The Geometry-Aware BEV Encoder provides metric spatial features in the BEV frame for grounding local targets under occlusion, while the Post-Fusion Affordance Decoder combines these features with the VLM output to predict a dense ego-centric BEV navigation affordance heatmap. To encourage structurally valid target prediction in traversable space and reduce sensitivity to imprecise target annotations, Stage~2 is trained with \textbf{Geodesic Target Region Supervision} (Section~\ref{sec:region_supervision}). At inference time, the final navigation target is obtained by taking the argmax of the predicted heatmap.

\subsection{Ego-Aligned Vision-Language Model}
\label{sec:vlm_3d}
The Ego-Aligned VLM provides instruction-conditioned ego-centric scene understanding from surround-view RGB inputs. It interprets spatial language in the agent frame and outputs a compact signal for BEV target prediction.

\subsubsection{Ego-Centric 3D Position Encoding}

To improve ego-centric scene understanding, we incorporate ego-centric 3D position information into visual tokens following LLaVA-3D~\cite{zhu2025llava} and SpatialVLA~\cite{qu2025spatialvla}, adapted to a surround-view setting. Given a 2D image patch token $v_i$ from the frozen vision transformer image encoder, we compute its depth-derived 3D position $p_i = (x_i, y_i, z_i)$ in the agent frame. A learnable embedding function $E_{3D}(\cdot)$, implemented as a lightweight two-layer multi-layer perceptron (MLP), maps $p_i$ to the visual feature dimension and is added to the corresponding visual token before the vision-to-language MLP projector:
\begin{equation}
\label{eq:3d_enc}
\tilde{v}_i = v_i + E_{3D}(p_i)
\end{equation}
Then, the MLP projector maps $\tilde{v}_i$ into the language model embedding space.

\subsubsection{Navigation Task Token Interface}

To obtain a single instruction-dependent signal for downstream Bird's-Eye-View prediction, we append a special token \texttt{[NAV]} to the prompt and use its final hidden state as a summary embedding following the common practice in vision-language robotic systems like TrackVLA~\cite{wang2025trackvla}, and the embedding is used as the vision-language input to the post-fusion affordance decoder.

\subsubsection{Stage-1 Auto-Derived Ego-Centric Instruction Tuning}

In Stage~1, we perform auto-derived ego-centric instruction tuning with a standard language modeling objective, optimizing the vision-to-language MLP projector, the ego-centric 3D position embedding $E_{3D}$, the language model’s low-rank adaptation (LoRA)~\cite{hu2022lora} parameters. Supervision is constructed automatically from the annotated target in the agent frame as coarse direction-and-range answers. Concretely, direction is discretized into eight 45$^\circ$ bins (e.g. Front, FrontLeft, etc.), and range is split into small or big by a fixed threshold $d_{range}$. These labels are expressed as short templated textual answers (e.g., ``Move towards the FrontLeft region with a small step.''), enabling the model to learn the ego-centric convention and integrate surround-view evidence.

In Stage~2, the trained model provides the \texttt{[NAV]} summary embedding while the full BEV affordance prediction is trained with geodesic target region supervision.

\subsection{Geometry-Aware Bird's-Eye View Encoder}
\label{sec:bev_path}

The Geometry-Aware BEV Encoder constructs an ego-centric BEV feature map $F_{\mathrm{BEV}}$ from two complementary sources: (i) dense image features projected to the ground plane using depth, camera calibration, and Bird's-Eye-View pooling; and (ii) depth geometry features produced by voxelizing depth points and encoding them with a 3D convolutional depth encoder based on SECOND~\cite{yan2018second}. Dense image features are extracted using a separate frozen vision backbone DINOv2~\cite{oquab2023dinov2}, distinct from the vision encoder inside the VLM, to preserve high-resolution detail for BEV projection.

We also compute an auxiliary BEV free-space cue $M$ from the current depth observation via ray casting, summarizing which cells are directly observed as free space. This cue is used to predict a per-cell gate $G \in [0,1]$ that controls the relative contribution of image features and geometry features. Concretely, the two BEV sources are mixed and projected as:
\begin{equation}
F_{\mathrm{BEV}} = \phi\Big(\big[(1-G)\odot F_{\mathrm{BEV}}^{\mathrm{Img}},\; M,\; G\odot F_{\mathrm{BEV}}^{\mathrm{Geom}}\big]\Big),
\end{equation}
where $F_{\mathrm{BEV}}^{\mathrm{Img}}$ denotes the depth-projected BEV image features, $F_{\mathrm{BEV}}^{\mathrm{Geom}}$ denotes the BEV geometry features from the depth encoder, $\odot$ is element-wise multiplication, $[\cdot]$ is channel-wise concatenation, and $\phi(\cdot)$ is a $1\times1$ projection.

\subsection{Post-Fusion Affordance Decoder}
\label{sec:post_fusion_decoder}

The Post-Fusion Affordance Decoder predicts a dense ego-centric BEV navigation affordance heatmap $\hat{A}$ by fusing the BEV feature map $F_{\mathrm{BEV}}$ with the compact embedding $F_{\mathrm{[NAV]}}$ produced by the Ego-Aligned VLM. We map $F_{\mathrm{[NAV]}}$ to a Bird's-Eye-View-aligned feature map via a convolutional upsampling projection to match the BEV grid, concatenate it with $F_{\mathrm{BEV}}$, and predict the BEV affordance heatmap with a standard BEV feature fusion module 
from BEVFusion~\cite{liu2023bevfusion}
 followed by convolutional layers.

\subsection{Geodesic Target Region Supervision}
\label{sec:region_supervision}

Point-only supervision provides weak guidance for dense BEV affordance prediction because it marks a single target location but does not explicitly indicate where not to predict. We adopt BEV target region supervision by aggregating depth observations from a small temporal window around the annotated target to obtain a local traversability estimate. This assumes reasonable depth quality and local pose consistency, and does not rely on simulator ground-truth maps.

Given an annotated target point $p^*$ on the BEV grid, the target region is defined as cells within a geodesic radius $r$:
\begin{equation}
\label{eq:region_sup}
R(p^*) = \{u \mid d_{\mathrm{geo}}(u, p^*) \le r\},
\end{equation}
where $d_{\mathrm{geo}}$ is the geodesic distance. Cells in $R(p^*)$ are treated as positives and all other cells as negatives, and we train with a binary cross-entropy loss between $\hat{A}$ and the target region mask.

\section{Experimental Setup}
\label{sec:experiment_setup}

We evaluate BEACON on language-conditioned local navigation target prediction, 
and additionally analyze performance on an occluded-target subset.
This section describes the experimental setup, including data construction in the Habitat simulator~\cite{puig2024habitat3} (Section~\ref{sec:data}), the compared baselines (Section~\ref{sec:baselines}), the evaluation metrics (Section~\ref{sec:metrics}), and implementation details (Section~\ref{sec:impl_details}). Results are presented in Section~\ref{sec:results_analysis}.

\subsection{Data Construction}
\label{sec:data}
We derive local navigation samples from Landmark-RxR~\cite{he2021landmark, ku2020room} by converting each instruction segment into \mbox{(start viewpoint, instruction, target)}, where the target is the segment endpoint viewpoint. At each start viewpoint, we render 4 surround-view RGB-D cameras ($448\times448$, $90^\circ$ FOV each). We restrict targets to a bounded local region ($\pm 6.4$\,m) and filter out samples outside this bound, requiring exploration beyond the local area (approximated by horizontal raycasts on the scene mesh), or with large height changes ($>0.5$\,m). The resulting split contains 70 scenes with 75K training samples and 12K unseen validation samples.

\textbf{Occluded-target subset.} We define an occluded-target subset using a depth-consistency test: the target is projected into each rendered view and marked occluded if its projected depth exceeds the rendered depth by more than $0.1$\,m in all views. Under this definition, 35.84\% / 34.37\% of train/validation samples are in this subset. To better reflect realistic occlusions from both scene structure and people, we also introduce non-interactive moving pedestrians, implemented with the simulator extension from Social-MP3D~\cite{gong2025cognition}. Pedestrian motion is randomized and collision-avoiding. 
The resulting subset has a slightly larger median target distance than the full validation set (3.12\,m vs.\ 2.32\,m).

\subsection{Baselines}
\label{sec:baselines}

We compare BEACON with three groups of methods: general-purpose VLM baselines, spatial-grounding VLM baselines evaluated with oracle-view selection, and a trained task-specific model. The general-purpose VLM baseline is ChatGPT-4o, which we prompt to output image-space points following the RoboRefer~\cite{zhouroborefer} prompting setup, and evaluate either on all four views jointly or with oracle-view selection. The spatial-grounding VLM baselines are RoboPoint~\cite{yuan2025robopoint} and RoboRefer, which include navigation-related target grounding as part of their capabilities. In our experiments, we use the RoboPoint-13B checkpoint and the largest publicly released RoboRefer checkpoint, denoted as RoboRefer-8B-SFT. RoboRefer-8B-SFT is our strongest open-source image-space baseline in this setting. Because they use a single-view image-space interface, we evaluate them with oracle-view selection. We additionally report RoboPoint-13B (best point) as a diagnostic upper bound on candidate selection, as RoboPoint outputs multiple point candidates per query. These methods output image-space predictions, so we evaluate them in zero-shot transfer rather than retraining them under our ground-plane target supervision, because occluded navigation targets do not have a well-defined image-space label in the current observation.
Finally, as the most straightforward supervised alternative, we train the same VLM with an MLP head to regress a single target point in BEV space from the images and instruction, testing whether BEACON’s gains can be explained by straightforward supervised adaptation alone.

\subsection{Evaluation Metrics}
\label{sec:metrics}
Following RoboPoint, we report thresholded target accuracy as the percentage of predicted points that fall within a target region. Because each sample provides a single annotated target point, we evaluate against a radius-$t$ region rather than exact point equality, reducing sensitivity to annotation imprecision and local endpoint ambiguity. We instantiate this in two ways: $\mathrm{GeoAcc}@t$ and $\mathrm{EucAcc}@t$ at $t\in\{0.5,1.0,1.5\}$\,m, where GeoAcc uses a geodesic target region of radius $t$ in traversable free space and EucAcc uses a Euclidean target region of radius $t$ on the ground plane. GeoAcc is the main metric because it reflects both localization and traversability, while EucAcc isolates spatial proximity even when a prediction falls inside static structure. We also report SIR (structural invalid rate), the fraction of predictions inside non-traversable static structure, to measure geometric validity directly. In the main tables, we report the average over thresholds, denoted by $\overline{\mathrm{GeoAcc}}$ and $\overline{\mathrm{EucAcc}}$.

\subsection{Implementation}
\label{sec:impl_details}

We train BEACON in two stages for one epoch each. Stage~1 uses learning rate $3\times10^{-5}$, with $d_{\mathrm{range}}=2.4$\,m as defined in Section~\ref{sec:vlm_3d}. Stage~2 uses base learning rate $2\times10^{-5}$, with a $5\times$ multiplier for the BEV encoder and the post-fusion decoder. We use InternVL2-2B~\cite{chen2024internvl} as the VLM throughout the experiments, optimizing the vision-to-language MLP projector, token embeddings, and the language model with LoRA (rank 16, alpha 256, dropout 0.05) while keeping the vision encoder frozen. All experiments run on a single NVIDIA A40 GPU with batch size 4 and gradient accumulation 2. The target geodesic radius $r$ is set to $1$\,m.

\begin{table*}[t]
\centering
\caption{Overall quantitative results on local navigation target prediction, comparing image-space baselines, straightforward trained alternative, and BEACON on the full validation set and occluded-target subset. Best results are shown in \textbf{bold}.}
\label{tab:overall_grouped_results}
\footnotesize
\setlength{\tabcolsep}{2.8pt}

\begin{tabular}{l c c | c c c c | c c c c}
\toprule
\multirow{2}{*}{\text{Method}} &
\multirow{2}{*}{\text{Input}} &
\multirow{2}{*}{\text{Output}} &
\multicolumn{4}{c|}{\textbf{Full Validation Set (\%)}} &
\multicolumn{4}{c}{\textbf{Occluded-Target Subset (\%)}} \\
& & & 
${\overline{\mathrm{GeoAcc}}}$$\uparrow$ &
${\overline{\mathrm{EucAcc}}}$$\uparrow$ &
\text{SIR}$\downarrow$ &
${\overline{\mathrm{GeoAcc}}}_{\mathrm{snap}}^{\dagger}\uparrow$ &
${\overline{\mathrm{GeoAcc}}}$$\uparrow$ &
${\overline{\mathrm{EucAcc}}}$$\uparrow$ &
\text{SIR}$\downarrow$ &
${\overline{\mathrm{GeoAcc}}}_{\mathrm{snap}}^{\dagger}\uparrow$ \\
\midrule

\multicolumn{11}{l}{\textit{General-purpose VLM baselines}} \\
\midrule
ChatGPT-4o~\cite{achiam2023gpt} & RGB & image point & 9.69 & 20.65 & 57.25 & 18.94 & 5.69 & 11.55 & 54.03 & 10.39 \\

ChatGPT-4o~\cite{achiam2023gpt} (oracle-view) & RGB & image point & 15.97 & 30.79 & 48.20 & 28.28 & 9.52 & 17.11 & 41.68 & 15.31 \\
\midrule
\multicolumn{11}{l}{\textit{Spatial-grounding VLM baselines with oracle-view selection}} \\
\midrule
RoboPoint-13B~\cite{yuan2025robopoint} & RGB & image point & 20.86 & 32.59 & 39.34 & 30.96 & 15.43 & 23.46 & 35.18 & 21.88 \\
RoboPoint-13B~\cite{yuan2025robopoint} (best point) & RGB & image point & 35.86 & 46.96 & 27.63 & 45.50 & 29.14 & 37.72 & 26.96 & 36.42 \\
RoboRefer-8B-SFT~\cite{zhouroborefer} & RGB-D & image point & 38.00 & 44.65 & 15.97 & 42.47 & 20.09 & 25.45 & 21.49 & 23.65 \\
\midrule

\multicolumn{11}{l}{\textit{Trained task-specific models}} \\
\midrule
VLM + point head & RGB & BEV point & 41.25 & 50.15 & 19.81 & 47.50 & 32.15 & 39.17 & 20.00 & 36.99 \\
\rowcolor{gray!15}
\text{BEACON (Ours)} & RGB-D & BEV heatmap & \textbf{57.72} & \textbf{60.17} & \textbf{2.13} & \textbf{58.50} & \textbf{42.83} & \textbf{45.36} & \textbf{2.60} & \textbf{43.56} \\
\bottomrule
\end{tabular}

\vspace{-2pt}
\begin{flushleft}
\footnotesize$^{\dagger}$ GeoAcc$_{\mathrm{snap}}$ is computed after snapping the prediction to the nearest oracle traversable cell as a diagnostic upper bound.\\
\end{flushleft}
\end{table*}

\section{Results}
\label{sec:results_analysis}

We analyze experiment results to answer three primary questions: 
\begin{itemize}
    \item  How does BEACON compare with image-space baselines and the most straightforward trained alternative on local navigation target prediction under occlusion? 
    \item To what extent does each proposed design choice contribute to accuracy and structural validity?
    \item What qualitative behaviors and failure modes does BEACON exhibit in challenging navigation cases? 
\end{itemize}

We answer these via quantitative results (Section~\ref{sec:main_results}) and qualitative analysis (Section~\ref{sec:qualitative}) respectively.

\subsection{Quantitative Results}
\label{sec:main_results}
Table~\ref{tab:overall_grouped_results} reports the main comparison against the baselines defined in Section~\ref{sec:baselines} on the full validation set and the occluded-target subset, while Table~\ref{tab:ablation_grid_template} analyzes the contribution of the Ego-Aligned VLM and BEV-space design choices with an ablation study. In Table~\ref{tab:ablation_grid_template}, removing both BEV Encoder and BEV Output gives an Ego-Aligned VLM with an MLP point head; removing only BEV Encoder gives an Ego-Aligned VLM with an MLP heatmap head; and removing only BEV Output gives an Ego-Aligned VLM whose output is updated by attending to BEV features through cross-attention before an MLP point head. Based on these results, we draw the following findings:

\begin{table}[t]
\centering
\scriptsize
\caption{Ablation study of key Ego-Aligned VLM and BEV-space design choices. Best results are shown in \textbf{bold}.}
\label{tab:ablation_grid_template}
\renewcommand{\arraystretch}{1.15}
\setlength{\tabcolsep}{3pt}

\begin{tabular}{c c c c | c | c c c}
\toprule
\multirow{2}{*}{\text{\shortstack{Stage 1\\Tuning}}} &
\multirow{2}{*}{\text{\shortstack{3D Pos.\\Enc.}}} &
\multirow{2}{*}{\text{\shortstack{BEV\\Encoder}}} &
\multirow{2}{*}{\text{\shortstack{BEV\\Output}}} &
\textbf{Val. (\%)} &
\multicolumn{3}{c}{\textbf{Occluded-Target Subset (\%)}} \\
& & & &
${\overline{\mathrm{GeoAcc}}}$$\uparrow$ &
${\overline{\mathrm{GeoAcc}}}$$\uparrow$ &
${\overline{\mathrm{EucAcc}}}$$\uparrow$ &
\text{SIR}$\downarrow$ \\
\midrule

\multicolumn{8}{l}{\textit{\textcolor{DarkBlue}{Ego-Aligned VLM design ablations}}} \\
\midrule
 &  & \checkmark  & \checkmark  & 54.76 & 40.06 & 42.49 & \textbf{2.37} \\
{\color{DarkBlue}\checkmark}  &   &  \checkmark &  \checkmark & 54.36 & 40.22 & 42.64 & 2.62 \\
  &  {\color{DarkBlue}\checkmark} &  \checkmark & \checkmark  & 53.59 & 37.93 & 40.31 & 2.50 \\
\midrule

\multicolumn{8}{l}{\textit{\textcolor{DarkGreen}{BEV-space design ablations}}} \\
\midrule
\checkmark  &  \checkmark &   &   & 48.40 & 37.26 & 43.82 & 16.53 \\
\checkmark  & \checkmark  & {\color{DarkGreen}\checkmark}  &   & 48.57 & 37.45 & 43.84 & 15.73 \\
\checkmark  &  \checkmark &   & {\color{DarkGreen}\checkmark}  & 52.80 & 36.97 & 42.01 & 11.08 \\
  
\midrule
\rowcolor{gray!15}
\checkmark & \checkmark  & \checkmark  & \checkmark  & \textbf{57.72} & \textbf{42.83} & \textbf{45.36} & {2.60} \\
\bottomrule
\end{tabular}
\end{table}

\noindent\textbf{Finding 1: BEACON substantially outperforms prior image-space baselines, especially under occlusion.} Table~\ref{tab:overall_grouped_results} shows that BEACON achieves the best results among all compared methods on both the full validation set and the occluded-target subset. This holds across both general-purpose VLM baselines and spatial-grounding VLM baselines. Compared with RoboRefer-8B-SFT, the state-of-the-art image-space baseline in our setting, BEACON improves occluded-subset $\overline{\mathrm{GeoAcc}}$ by 22.74 percentage points and reduces SIR from 21.49\% to 2.60\%. Together, these results show a consistent gap between BEACON and prior image-space baselines in both target accuracy and structural validity, especially under occlusion.

\noindent\textbf{Finding 2: Straightforward supervised adaptation alone is insufficient.} Table~\ref{tab:overall_grouped_results} shows that training the same VLM with an MLP point head improves over prior image-space baselines, confirming that task-specific supervision is beneficial. However, the gain remains limited: on the full validation set, its $\overline{\mathrm{GeoAcc}}$ is only 3.25 points higher than RoboRefer-8B, and it still remains clearly below BEACON on both accuracy and structural validity. Table~\ref{tab:ablation_grid_template} further shows that removing any major proposed component leads to a noticeable drop in performance. Together, these results indicate that BEACON’s gains do not come from supervised adaptation alone, but from the combined effect of its proposed design choices.

\noindent\textbf{Finding 3: BEACON’s gains are not just from post-hoc snapping.} Table~\ref{tab:overall_grouped_results} shows that BEACON improves $\overline{\mathrm{EucAcc}}$ and $\overline{\mathrm{GeoAcc}}_{\mathrm{snap}}$ in addition to $\overline{\mathrm{GeoAcc}}$, so its gains are not explained only by producing fewer invalid predictions and relying on snapping as a post-hoc correction. Table~\ref{tab:ablation_grid_template} further shows that the Ego-Aligned VLM design improves $\overline{\mathrm{EucAcc}}$ on the occluded-target subset, indicating better language-conditioned target prediction even under a metric that does not enforce structural validity. Notably, ego-centric 3D position encoding alone does not consistently help, and only becomes beneficial when combined with Stage-1 ego-centric instruction tuning, which further shows that the gain comes from the coordinated design of our Ego-Aligned VLM, rather than from adding 3D positional information in isolation.

\noindent\textbf{Finding 4: BEACON yields drastically lower non-traversable predictions.} Table~\ref{tab:overall_grouped_results} shows that BEACON achieves a drastically lower SIR on both the full validation set and the occluded-target subset (2.13 and 2.60, respectively), indicating that its predicted targets rarely fall inside non-traversable static structure. Table~\ref{tab:ablation_grid_template} supports that this improvement comes from BEV-space modeling: removing BEV components sharply increases SIR on the occluded-target subset (11.08--16.53). Notably, the lowest SIR is achieved only when both BEV Encoder and BEV Output are enabled, consistent with our design choice of combining BEV geometric features with a BEV-space affordance output.

\begin{table}[h]
\centering
\small
\caption{Ablation study on $F_{\mathrm{BEV}}$ components.}
\label{tab:bev_components}
\setlength{\tabcolsep}{4pt}
\renewcommand{\arraystretch}{1.15}

\begin{tabular}{c c c | c c}
\toprule
\multirow{2}{*}{$F_{\mathrm{BEV}}^{\mathrm{Img}}$} &
\multirow{2}{*}{$F_{\mathrm{BEV}}^{\mathrm{Geom}}$} &
\multirow{2}{*}{$G$} &
\textbf{Full Val. (\%)} &
\textbf{Occ. Subset (\%)} \\
& & &
${\overline{\mathrm{GeoAcc}}}$$\uparrow$ / \text{SIR}$\downarrow$ &
${\overline{\mathrm{GeoAcc}}}$$\uparrow$ / \text{SIR}$\downarrow$ \\
\midrule
$\checkmark$ &              &              & 55.99 / 3.22  & 41.52 / 3.77  \\
             & $\checkmark$ &              & 51.67 / 9.54  & 36.93 / 12.51 \\
$\checkmark$ & $\checkmark$ &              & 56.96 / \textbf{2.12} & 42.34 / 2.62  \\
\rowcolor{gray!15}
$\checkmark$ & $\checkmark$ & $\checkmark$ & \textbf{57.72} / 2.13 & \textbf{42.83} / \textbf{2.60} \\
\bottomrule
\end{tabular}
\end{table}

\textbf{BEV feature component ablation.}
Table~\ref{tab:bev_components} studies the BEV feature construction in Section~\ref{sec:bev_path} by ablating the image feature branch $F_{\mathrm{BEV}}^{\mathrm{Img}}$, the geometry feature branch $F_{\mathrm{BEV}}^{\mathrm{Geom}}$, and the learned gate $G$. Using only $F_{\mathrm{BEV}}^{\mathrm{Img}}$ already gives strong performance, while using only $F_{\mathrm{BEV}}^{\mathrm{Geom}}$ substantially lowers $\overline{\mathrm{GeoAcc}}$ and increases SIR. Combining the two branches improves both accuracy and validity, showing their complementarity. Adding the gate gives a further gain in $\overline{\mathrm{GeoAcc}}$ while preserving very low SIR, supporting the design of the Geometry-Aware BEV Encoder.

\subsection{Qualitative Analysis}
\label{sec:qualitative}
\begin{figure}[t]
    \centering

    \begin{subfigure}{\linewidth}
        \centering
        \includegraphics[width=\linewidth]{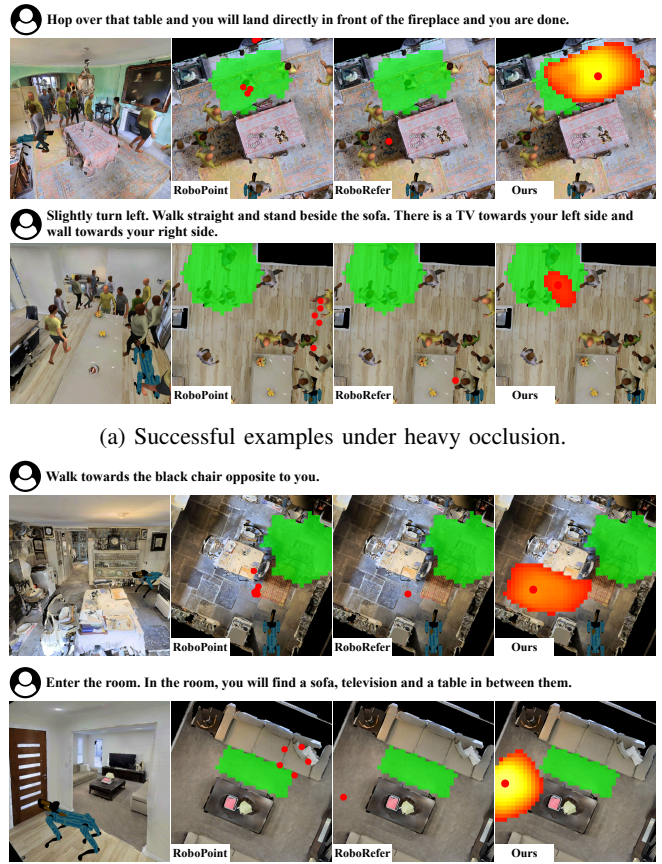}
        \caption{Successful examples under heavy occlusion.}
        \label{fig:working_analysis}
    \end{subfigure}

    \vspace{0.1cm}

    \begin{subfigure}{\linewidth}
        \centering
        \includegraphics[width=\linewidth]{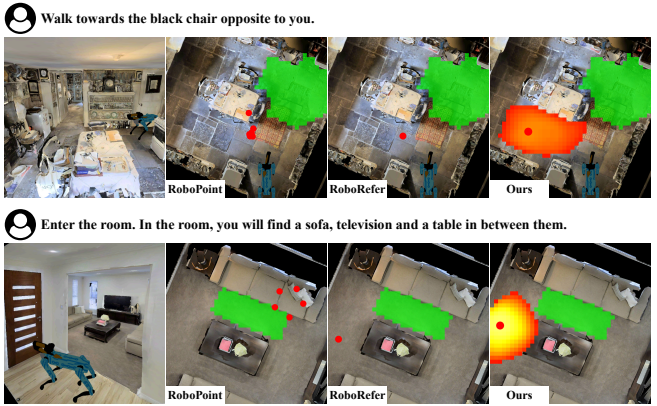}
        \caption{Failures due to landmark confusion or instruction ambiguity.}
        \label{fig:failure_cases}
    \end{subfigure}

    \caption{Qualitative examples of language-conditioned navigation affordance prediction, comparing BEACON’s BEV affordance predictions with the image-space baselines RoboPoint~\cite{yuan2025robopoint} and RoboRefer~\cite{zhouroborefer}. Target regions are shown in \textcolor{green}{green}.}
    \label{fig:qualitative_analysis}
    \vspace{-2em}
\end{figure}

Figure~\ref{fig:qualitative_analysis} compares BEACON’s BEV affordance predictions with the image-space baselines RoboPoint and RoboRefer. For readability, the heatmap overlay is thresholded at 0.40 so that only high-confidence regions are shown. The top two rows show successful examples under heavy occlusion, while the bottom two rows illustrate representative failure cases. Here, successful means that the selected target lies inside the 1\,m geodesic target region.

\textbf{Affordance prediction under heavy occlusion.} In the first successful example shown in Figure~\ref{fig:working_analysis}, part of the referred structure is visible, but the target-side free space is heavily occluded; BEACON concentrates affordance in the feasible gap and selects a target inside the target region. In the second example, the relevant landmarks are not directly visible and the observation provides mainly layout cues; BEACON still assigns probability mass toward the correct direction and rough location, whereas the image-space baselines fail without a directly visible grounding cue.

\textbf{Uncertainty representation and structural validity.} Our affordance map explicitly represents uncertainty as a spatial distribution over candidate free-space goals, while remaining anchored to traversable geometry. In the first successful example and the first failure case, probability mass follows feasible corridors around furniture rather than spreading into walls or obstacles, so the selected target is less likely to fall inside static structure. This behavior is encouraged by the geodesic region supervision, which provides explicit negatives on infeasible regions and suppresses non-traversable areas even when the semantic prediction is imperfect, consistent with the low SIR observed quantitatively. By contrast, image-space baselines do not explicitly model free-space feasibility: RoboRefer tends to select conservative visible-floor points that miss occluded targets, while RoboPoint may predict semantically relevant pixels whose depth projection is not traversable.

\textbf{Failure cases.} The two rows in Figure~\ref{fig:failure_cases} summarize two common failure modes. In the first row, the model confuses the referred landmark or relation (e.g., which chair is black and which one is ``opposite''), yielding a coherent but misplaced affordance peak. In the second row, the instruction is underspecified about how far to proceed after entering the room; the prediction corresponds to a plausible stopping region but exhibits an ambiguity-induced mismatch with the single annotated endpoint.

\section{Conclusion}

In this work, we propose BEACON, a VLM-based BEV affordance predictor for local navigation target prediction conditioned on an open-vocabulary instruction. In unseen environments in the Habitat simulator, BEACON shows consistent gains over prior image-space baselines, with the largest improvements on the occluded-target subset. While image-space baselines struggle with occluded cues or targets, BEACON outputs an ego-centric BEV affordance heatmap that yields more accurate targets and substantially fewer non-traversable predictions. These improvements are not simply the result of adding task-specific supervision, nor are they explained solely by post-hoc snapping to free space; instead, BEACON improves both Euclidean target accuracy and traversable-target validity. Extensive ablations further validate the importance of ego-aligned 3D cues and BEV-space design choices.

While BEACON demonstrates strong results in simulation, evaluating it on real-world surround-view RGB-D data with matched instruction segments is an important next step. Looking ahead, incorporating more explicit compositional grounding of intermediate entities and relations, together with process-level supervision, may further improve multi-step spatial reasoning.

\bibliographystyle{IEEEtran}  %
\bibliography{new_refs}

\end{document}